\documentclass[conference]{IEEEtran}
\usepackage{cite}
\usepackage{amsmath,amssymb,amsfonts}
\usepackage{algorithmic}
\usepackage{graphicx}
\usepackage{textcomp}
\usepackage{xcolor}
\def\BibTeX{{\rm B\kern-.05em{\sc i\kern-.025em b}\kern-.08em
    T\kern-.1667em\lower.7ex\hbox{E}\kern-.125emX}}
    
\usepackage{varioref}
\usepackage[dvipsnames]{xcolor}
\usepackage{graphicx}
\usepackage[T1]{fontenc}
\usepackage[utf8]{inputenc}\DeclareUnicodeCharacter{2212}{-}
\usepackage{url}
\usepackage{pgf}
\usepackage{etoolbox}
\usepackage{xstring}
\usepackage{tcolorbox}
\usepackage{hyperref}
\usepackage{algorithm}
\usepackage{listings}
\usepackage{booktabs}
\usepackage{siunitx}
\lstdefinestyle{pytorch}{
    backgroundcolor=\color{white},
    basicstyle=\ttfamily\footnotesize,
    numbers=left,
    numberstyle=\tiny\color{Gray},
    stepnumber=1,
    numbersep=5pt,
    xleftmargin=1.5em,
    framexleftmargin=1.5em,
    keywordstyle=\color{Blue},
    commentstyle=\color{OliveGreen},
    stringstyle=\color{BrickRed},
    breaklines=true,
    columns=fullflexible,
    language=Python,
    morekeywords={with, torch, no_grad},
    frame=lines,
}

\usepackage{tikz}
\usepackage{rotating} 
\usepackage[tikz]{ocgx2}
\usepackage{pgfplotstable}
\usepackage{pgfplots}
\usepackage{tabularx}
\usetikzlibrary{arrows,fit,positioning,shapes.symbols,chains,calc,quotes,arrows.meta,shapes.geometric,decorations.markings}
\pgfplotsset{width=4cm,compat=1.9}
\usepackage{multirow}

\newcounter{example}[section]

\title{Randomly Initialized Networks Can Learn from Peer-to-Peer Consensus}

\author {
\IEEEauthorblockN{Esteban Rodríguez-Betancourt}
\IEEEauthorblockA{Posgrado en Computación e Informática\\
Universidad de Costa Rica\\
esteban.rodriguezbetancourt@ucr.ac.cr}

\and

\IEEEauthorblockN{Edgar Casasola-Murillo}
\IEEEauthorblockA{
Escuela de Ciencias de la Computación\\
Universidad de Costa Rica\\
edgar.casasola@ucr.ac.cr}
}

\begin{document}



\IEEEoverridecommandlockouts


\maketitle


\newtcolorbox{mycolorbox}[3][]
{
  colframe = #2!25,
  colback  = #2!10,
  coltitle = #2!20!black,  
  title    = {#3},
  #1,
}

\IEEEpubidadjcol

\begin{abstract}
    \noindent
In self-supervised learning, self-distilled methods have shown impressive performance, learning representations useful for downstream tasks and even displaying emergent properties. However, state-of-the-art methods usually rely on ensembles of complex mechanisms, with many design choices that are empirically motivated and not well understood.

In this work, we explore the role of self-distillation within learning dynamics. Specifically, we isolate the effect of self-distillation by training a group of randomly initialized networks, removing all other common components such as projectors, predictors, and even pretext tasks. Our findings show that even this minimal setup can lead to learned representations with non-trivial improvements over a random baseline on downstream tasks. We also demonstrate how this effect varies with different hyperparameters and present a short analysis of what is being learned by the models under this setup.
\end{abstract}


\newcommand{\figref}[1]{\figurename{} \ref{#1}}



\section{Introduction}
Representation Learning consists in learning useful, general purpose data representations, that capture the structure and semantics of the input space \cite{bengio2013representation,goodfellow2016deeplearning}. Self-supervised learning (SSL) methods have demonstrated that they are well suited to learn useful representations, bypassing the need for labeled datasets. These methods have been applied successfully to diverse domains such as text and vision representations.

While self-supervised learning methods are undeniably successful, they usually rely on components that are not trivial to devise. For instance, a family of methods such as BYOL \cite{grill2020} and DINO \cite{caron2021emerging} rely on self-distilling representations from a randomly initialized network called teacher, while this teacher is updated using EMA mechanism. While the mechanisms clearly works, we lack a theoretical understanding of how they work \cite{wang2021towards}.

In this work, we investigate the effects on learning representations of self-distillation from a group of untrained networks. We observe non-trivial improvements in downstream tasks such as CIFAR-10 classification. However, those gains are dependent of the learning rate and model architecture.

In order to isolate any possible learning to just the self-distillation dynamics, we opted to strip away many components usually found in self-distillation mechanisms, such as pretext tasks that produce a view asymmetry, loss function mutations, student/teacher asymmetry, projector/predictor layers.

Even after removing all those components, we were able to get learned representations that were better than our random baseline, for downstream tasks such as CIFAR-10 \cite{krizhevsky2009learning} classification. Additionally, this learning mechanism remains stable and avoids representational collapse.

Under this simple setup, we studied the effect of several hyperparameters, such as learning rate, loss function, number of peers and teachers and differences based on network architecture. Additionally, we did an early exploration of what may have been learned by the networks under this setup.

\section{Concepts and Related work}\label{sec:RelatedWork}
\subsection{Representation Learning and Self-Supervision}

Representation learning aims to extract features from data that capture semantic structure and generalize well to downstream tasks \cite{bengio2013representation, goodfellow2016deeplearning}. In the absence of labeled data, self-supervised learning (SSL) has become a widely used strategy, particularly in domains such as computer vision and natural language processing. SSL methods learn from the data itself by designing pretext tasks or objectives that encourage the model to encode meaningful information. Notable examples include SimCLR with contrastive losses \cite{chen2020simclr}, and Barlow Twins for redundancy reduction \cite{zbontar2021barlow}.

\subsection{Self-Distillation in SSL}

A successful branch of SSL focuses on self-distillation, where one model is trained to predict the output of another. In BYOL \cite{grill2020} and SimSiam \cite{chen2021simsiam}, a student network learns to align its representations with those of a teacher network that receives a different view of the same input. These methods typically rely on architectural asymmetry, including a predictor head and a stop-gradient operation, to avoid trivial solutions. DINO \cite{caron2021emerging} replaces the predictor with a softmax-based loss that uses centering and sharpening to prevent collapse, and employs an EMA-updated teacher to stabilize training.

\subsection{Preventing Collapse in Non-Contrastive SSL}

A typical issue in self-distilled self-supervised learning models is that the network can end up collapsing most of the inputs into the same embedding. This issue has been observed in multiples methods and many different solutions have been proposed. In SimSiam \cite{chen2021simsiam}, authors noticed that stopping the gradient from to the ``teacher'' side was fundamental to stop the collapse. DINO \cite{caron2021emerging} relied on loss centering and scaling to prevent the collapse into the same representation. Other authors opt to tackle directly the collapse. For instance, SimDINO \cite{wu2025simplifyingdinocodingrate} introduced an explicit expansion component into the loss based on coding rate regularization.

\subsection{Random Teachers and Self-Distillation without Training}

Recent work by Sarnthein et al. \cite{sarnthein2023random} explores a surprising setting where a fixed randomly initialized teacher can still guide a student to learn transferable representations. Their results show that even learning to mimic a frozen untrained teacher is sufficient for learning meaningful features. They also highlight how proximity between teacher and student initialization affects learning dynamics. This behavior has been observed previously, for example, it is mentioned as a core motivation behind BYOL \cite{grill2020}.

Like Sarnthein et al., we explore the question of what is learned under a minimal setup without pretext tasks and other mechanisms of complete SSL methods. However, in this work, rather than using a fixed teacher, we use a peer-to-peer group of randomly initialized networks. So, we can study the effect of learning from a moving randomly initialized network, without introducing the asymmetric effect of an EMA teacher.


\section{Methodology}\label{sec:Methodology}
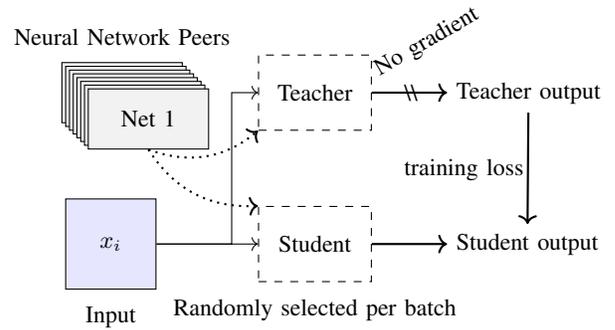
\begin{figure}[htbp]
    \centering
\begin{tikzpicture}[
    every node/.style={font=\small},
    node distance=1.0cm and 1.0cm,
    net/.style={rectangle, draw, minimum width=1.6cm, minimum height=0.8cm, align=center, fill=gray!10},
    temp/.style={rectangle, draw, dashed, minimum width=1.5cm, minimum height=1cm, align=center},
    img/.style={rectangle, draw, minimum width=1.2cm, minimum height=1.2cm, fill=blue!10},
    loss/.style={diamond, draw, fill=red!10, inner sep=1pt},
    grad/.style={->, thick},
    nograd/.style={->, thick},
    stop/.style={rectangle, draw, minimum width=0.8cm, minimum height=0.5cm, fill=gray!20}
]

\foreach \i in {0,...,7} {
    \node[net, yshift=-0.05cm*\i, xshift=0.05cm*\i] (net\i) at (0,0) {Net \the\numexpr 7-\i+1};
}
\node[above=0.1cm of net0] {Neural Network Peers};

\node[img, below=0.65cm of net7, xshift=-0.5cm] (img) {$x_i$};
\node[below=0.1 of img]{Input};

\node[temp, right=1cm of net0] (teacher) {Teacher};
\node[temp, below=1cm of teacher] (student) {Student};

\draw[->, dotted, thick, bend right=30] (net7.south) to (teacher.south west);
\draw[->, dotted, thick, bend right=30] (net7.south) to (student.north west);

\draw[->] (img.east) -- ++(1.0,0) |- (teacher.west);
\draw[->] (img.east) -- ++(1.0,0) |- (student.west);

\node[right=1cm of teacher] (tout) {Teacher output};
\node[right=1cm of student] (sout) {Student output};

\draw[nograd] (teacher.east) -- (tout);
\node[scale=1](nogradsym) at ($(teacher.east)!0.25!(tout)$) {\textbackslash{}\textbackslash{}};
\node[above=0.01cm of nogradsym,rotate=30,xshift=0.4cm]{No gradient};

\draw[grad] (student.east) -- (sout);

\draw[->, thick] (tout) -- node[below, xshift=-0.85cm, yshift=0.25cm] {training loss} (sout);

\node[below=0.1cm of student] {Randomly selected per batch};

\end{tikzpicture}
    \caption{Schematic of \emph{DINOHerd} setup with a single teacher and student chosen at random per batch.}
    \label{fig:herdschematic}
\end{figure}

Our setup is inspired by DINO \cite{caron2021emerging}, however, we stripped away most of the components to focus exclusively on the effects on learning of self distillation. In \emph{DINOHerd}, we start with a group of untrained, randomly initialized neural networks. The specific architecture of the network is not relevant, as long as it is reasonable for the task at hand and it returns embeddings we can compare with a loss function. For each batch, we randomly choose one student and $T$ teachers. For each teacher and student, we generate the corresponding output for the given batch. We stress each network sees exactly the same view of the data, the teachers do not receive a different augmentation than the student. Finally, we use a loss function to determine the difference between the student and the teachers, making sure to only back propagate the error through the student, not the teachers. PyTorch pseudocode is shown in Listing \ref{alg:dinoherd} and a schematic of our training setup is shown at Figure \ref{fig:herdschematic}.

\begin{figure}[htbp]
\centering
\begin{lstlisting}[style=pytorch, caption={PyTorch pseudocode for DINOHerd}, label={alg:dinoherd}]
# N: Number of peers
# T: Number of teachers
peers = [Network() for _ in range(N)]
optimizers = [make_optimizer(p) for p in peers]

for x in dataloader:
    # Sample student and teachers from pool
    idxs = random.sample(range(len(peers)), T + 1)
    student_idx = idxs[0]

    student = peers[student_idx]
    teachers = [peers[i] for i in idxs[1:]]

    # Forward pass for student
    student_out = student(x)

    # Forward pass for teachers (frozen)
    with torch.no_grad():
        teacher_outs = [teacher(x)
            for teacher in teachers]

    # Average loss between student/teachers
    losses = [loss_fn(student_out, teacher_out) for teacher_out in teacher_outs]
    loss = torch.mean(torch.stack(losses))

    # Backprop through student only
    optimizers[student_idx].zero_grad()
    loss.backward()
    optimizers[student_idx].step()
\end{lstlisting}
\end{figure}

\subsection{Teacher dynamics}
In our framework, the teacher role is assigned randomly to $N$ networks. This assignment lasts for only a single batch. This creates an environment where no peer has a fixed special role during the whole training: each peer will be a student or a teacher at any moment during the training. This dynamic simplifies our setup, as we do not need to implement any different logic for updating the weights of the teacher. As is typical with other similar methods, we use a stop-gradient operator on the teacher branch during backpropagation; only the current student weights are updated per batch.

We explored the effect of choosing more than one teacher per batch. The results of those experiments are shown in Subsection \ref{subsec:NumberOfTeachers}.

\subsection{Loss function and learning rate}
During development, we experimented with traditional loss functions such as cosine loss and mean squared error (MSE). However, we observed that lower learning rates consistently improved the stability and quality of the learned representations. A lower learning rate is equivalent to scaling the gradient, which we show as follows:

\paragraph*{Learning rate and gradient scaling equivalence}
Consider a learning rate defined as \( \eta = \alpha \cdot C \). Let \( \theta_t \) be the weights of a neural network at training step \( t \). The SGD update rule is 
\(
\theta_{t+1} = \theta_t - \eta \cdot \nabla \mathcal{L}(\theta_t)
\) .
Substituting the learning rate \( \eta \), we obtain:
\(
\theta_{t+1} = \theta_t - \alpha \cdot C \cdot \nabla \mathcal{L}(\theta_t)
\) . Then, by the linearity of the gradient operator with respect to scalar multiplication, we get:
\(
\theta_{t+1} = \theta_t - C \cdot \nabla \left( \alpha \cdot \mathcal{L}(\theta_t) \right)
\) .

Since the learning rate effectively acts as a multiplicative scaling factor on the loss, we began to reinterpret it as a form of implicit temperature control on the loss landscape (conceptually akin to the temperature scaling used in DINO \cite{caron2021emerging}). A lower loss scale reduces the influence of smaller gradients, allowing only the most salient error signals to drive updates.

This observation motivated the design of our \emph{salient loss}, which focuses training on the single most divergent feature dimension per sample. Rather taking all the dimensions into account, our \emph{salient loss} only considers the dimension with the maximum distance ($max((A-B)^2)$). Interestingly, introducing this custom loss did not changed much the learning dynamics, as explained in subsection \ref{subsec:lossfunc}.

Although the salient loss provides a focused gradient signal, we found that increasing the learning rate led to worse representations. The best results were obtained using small learning rates with the Adam or AdamW optimizers in combination with the salient loss. This suggests that while saliency helps guide learning, it must be coupled with gradual updates to maintain alignment across the peer networks.

\subsection{Training}

\begin{figure}[htpb]
    \centering

\begin{tikzpicture}[
  node distance=0.4cm and 0.4cm,
  box/.style={draw, minimum width=0.5cm, minimum height=2cm, align=center},
  conv/.style={box, fill=blue!20},
  bn/.style={box, fill=yellow!20},
  act/.style={box, fill=red!20},
  pool/.style={box, fill=lime!50},
  arrow/.style={-Latex, thick},
  font=\footnotesize
]

\node[conv] (conv1) {\rotatebox{90}{Conv2D 3 to 64}};
\node[bn, right=of conv1] (bn1) {\rotatebox{90}{BatchNorm}};
\node[act, right=of bn1] (act1) {\rotatebox{90}{LeakyReLU}};
\node[conv, right=of act1] (conv2) {\rotatebox{90}{Conv2D 64 to 128}};
\node[bn, right=of conv2] (bn2) {\rotatebox{90}{BatchNorm}};
\node[act, right=of bn2] (act2) {\rotatebox{90}{LeakyReLU}};
\node[pool, right=of act2] (pool1) {\rotatebox{90}{MaxPool2D ($2\times2$)}};

\node[conv, below=of conv1] (conv3) {\rotatebox{90}{Conv2D 128 to 256}};
\node[bn, right=of conv3] (bn3) {\rotatebox{90}{BatchNorm}};
\node[act, right=of bn3] (act3) {\rotatebox{90}{LeakyReLU}};
\node[conv, right=of act3] (conv4) {\rotatebox{90}{Conv2D 256 to 256}};
\node[bn, right=of conv4] (bn4) {\rotatebox{90}{BatchNorm}};
\node[act, right=of bn4] (act4) {\rotatebox{90}{LeakyReLU}};
\node[pool, right=of act4] (pool2) {\rotatebox{90}{MaxPool2D ($2\times2$)}};

\foreach \a/\b in {conv1/bn1, bn1/act1, act1/conv2, conv2/bn2, bn2/act2, act2/pool1, conv3/bn3, bn3/act3, act3/conv4, conv4/bn4, bn4/act4, act4/pool2} {
  \draw[arrow] (\a) -- (\b);
}

\draw[arrow] (pool1.east) -- ++(0.5,0) -- ++(0, -1.25) -- ++(-7, 0) -- ++(0, -1.26) -- (conv3);

\node[left=0.4cm of conv1] (input) {\rotatebox{90}{Input Image}};
\node[right=0.4cm of pool2] (output) {\rotatebox{90}{Feature Map}};

\draw[arrow] (input) -- (conv1);
\draw[arrow] (pool2) -- (output);

\end{tikzpicture}

    \caption{Diagram of the neural network used for learning vision representations}
    \label{fig:visioncnn}
\end{figure}
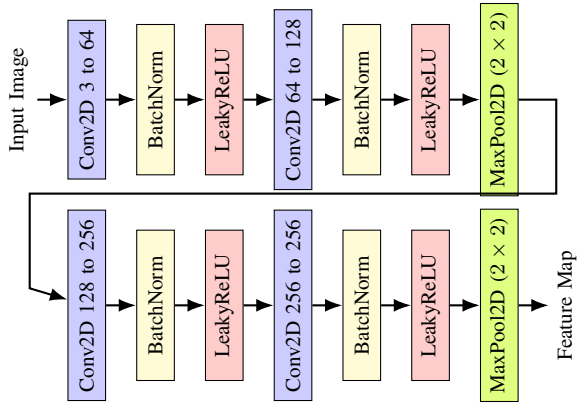

For both training and evaluating our models, we used CIFAR-10 dataset \cite{krizhevsky2009learning}. It consists of 60000 32x32 color images of ten different classes: airplane, automobile, bird, cat, deer, dog, frog, horse, ship and truck. The only augmentation applied to the dataset was a 50\% probability horizontal flip. Unlike other self-supervised and self-distilled setups, in our case the teacher and the student always receive exactly the same view of the data. We used a simple convolutional neural network, as shown in Figure \ref{fig:visioncnn}. For evaluation, we trained a KNN (K=5), a linear probe and a MLP on top of the frozen backbone.

\section{Main Results}\label{sec:Results}
\subsection{Vision Results}
We trained a single vision model on CIFAR-10 dataset \cite{krizhevsky2009learning} for 10 epochs using a learning rate of 1e-8 and 16 peers. The training took a total of 2:14 minutes using a batch size of 512, being trained on a single NVIDIA GeForce RTX 3060 with 12GB of VRAM. The resulting model had a file size of 3.9MB.

\begin{table}[htbp]
  \centering
  \caption{CIFAR-10 classification results (16 peers, 10 epochs).}
  \label{tab:vision_results}
  \begin{tabular}{l S S S}
    \toprule
    \textbf{Network} & {\textbf{KNN (K=5) F1 (\%)}} & {\textbf{Linear F1 (\%)}} & {\textbf{MLP F1 (\%)}} \\
    \midrule
    Baseline & 37.22 & 16.97 & 15.71 \\
    Trained  & 41.58 & 43.43 & 49.26 \\
    \bottomrule
  \end{tabular}
\end{table}

\begin{figure}[htbp]
    \centering
    \includegraphics[width=1\linewidth]{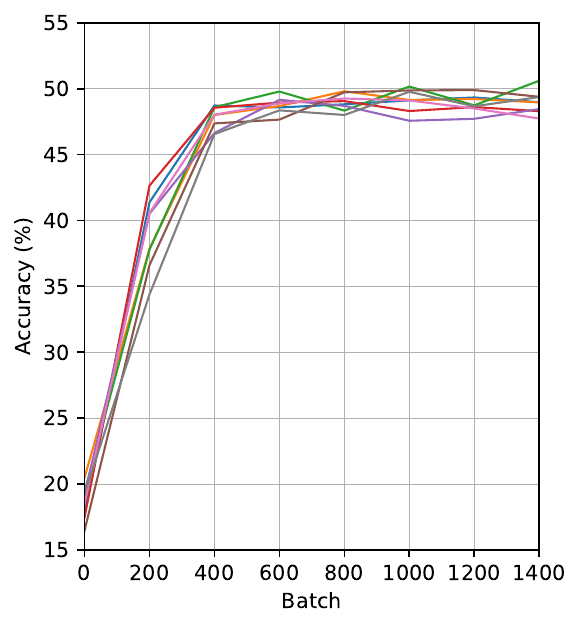}\caption{Comparison of linear probe accuracy of 8 peers on CIFAR-10.}
    \label{fig:groupdynamics}
\end{figure}

After training, we evaluated the learned representations using a KNN classifier ($K=5$), a linear probe and a small MLP. Both the linear probe and the MLP were trained for 20 epochs using SGD and a learning rate of 0.01. Table \ref{tab:vision_results} summarizes the results compared to an untrained baseline. As the table shows, the training process improved the classification results, specially when using linear or MLP probes.

To further understand the learning dynamics within the peers, we trained a smaller setup for a single epoch, with just eight randomly initialized models, using a learning rate of 1e-8 and a batch size of 32. Then, every 50 batches we trained a linear probe on CIFAR-10 to evaluate the quality of the learned features. As shown in Figure \ref{fig:groupdynamics}, the untrained networks start with low quality representations (mostly below $20\%$ linear probe accuracy) and in around 500 batches most networks stabilized. Also, the plot shows that the learning dynamics improves all the networks consistently, even if CIFAR-10 classification is never used as a training objective.

\section{Hyperparameter Exploration}\label{sec:Ablations}
In this section, we will describe the results of performing several variations on our training framework. All the ablation studies were done using our vision model trained on CIFAR-10.

\subsection{Effect of Number of Peers and Teachers Count}\label{subsec:NumberOfTeachers}
\begin{figure}[htbp]
    \centering
    \includegraphics[width=1\linewidth]{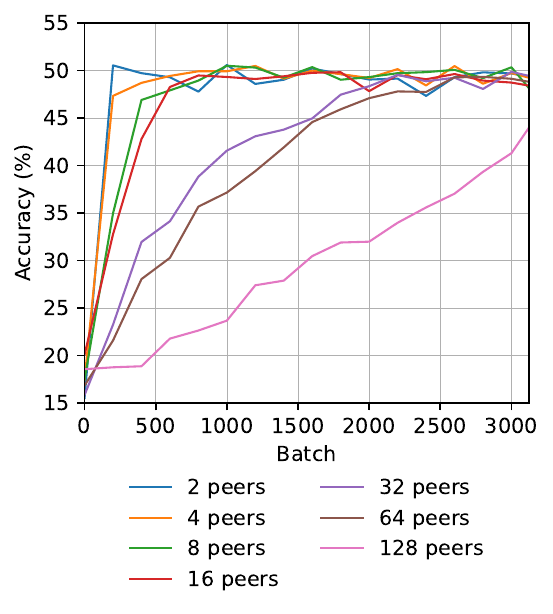}
    \caption{Comparison of linear probe accuracy varying the number of peers.}
    \label{fig:num-peers}
\end{figure}

\begin{figure}[htbp]
    \centering
    \includegraphics[width=1\linewidth]{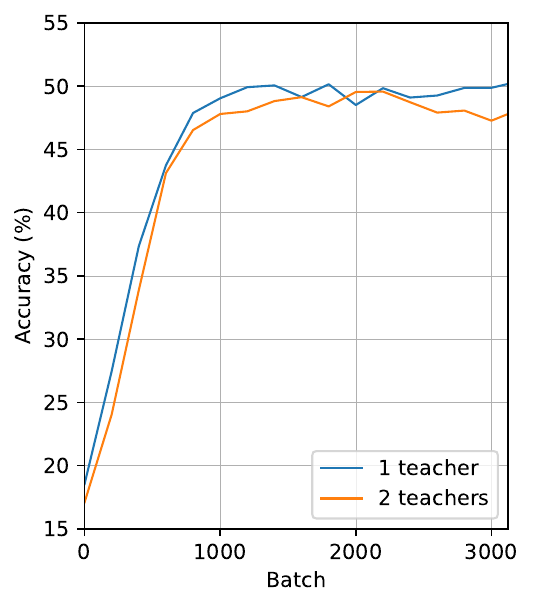}
    \caption{Comparison of accuracy varying the number of teachers.}
    \label{fig:num-teachers}
\end{figure}

When evaluating the impact of the number of peers and teachers on the learned representations, we found that, aside from making the learning slower, increasing the number of peers produce eventually similar results, as shown in Figure \ref{fig:num-peers}. Similarly, we did not find important differences between using one or two teachers, as shown in Figure \ref{fig:num-teachers}. For both cases, the accuracy was evaluated every 200 batches for two epochs, using salient loss, learning rate of $1e-8$ and a batch size of 32. For Figure \ref{fig:num-teachers} 16 peers were used.

\subsection{Individual vs Ensemble Performance}
\begin{figure}[htbp]
    \centering
    \includegraphics[width=1\linewidth]{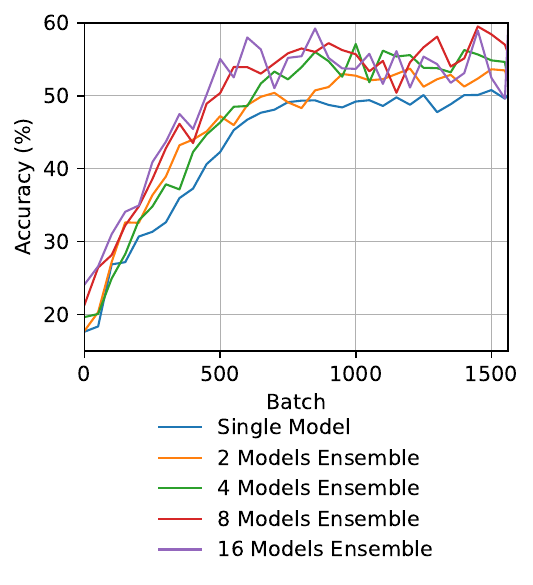}
    \caption{Comparison of the ensemble of a varying number of peer outputs.}
    \label{fig:swarm-results}
\end{figure}

When using a linear probe over the concatenated outputs of multiple peers, we observed a slight improvement of the downstream task performance of the models. As shown in the Figure \ref{fig:swarm-results}, the single model has lower accuracy than the ensembles of two or more models. This demonstrate that each peer is learning different features, that could be useful for classification. However, there are also diminishing gains on adding additional models outputs to the concatenated embeddings. In this case, the accuracy was evaluated evaluated on CIFAR-10 every 50 batches, with batch size 32 and learning rate 1e-8. In all cases, a 16 peers setup was used.

\subsection{Impact of the Loss Function and Network Architecture}\label{subsec:lossfunc}
\begin{figure}[htbp]
    \centering
    \includegraphics[width=1\linewidth]{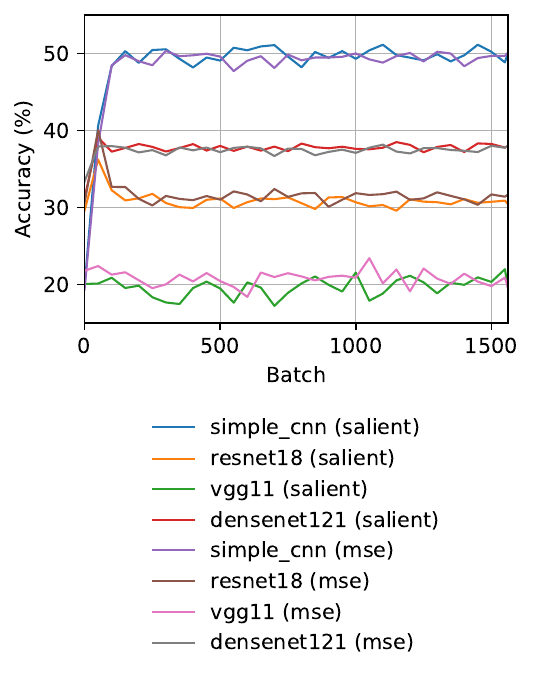}
    \caption{Comparison of linear probe accuracy varying network architecture and loss function.}
    \label{fig:loss-and-network}
\end{figure}

We compared the learning dynamics of different loss functions (MSE, salient loss) and network architectures (Simple CNN, ResNet18 \cite{kaiming2016resnet}, VGG11 \cite{liu2015vgg11}, DenseNet121 \cite{gao2017densenet121}) on CIFAR-10 using our setup. In this case, we used two peers, one teacher, a learning rate of 1e-8, and a batch size of 32, and evaluated CIFAR-10 performance every 50 batches. As shown in Figure \ref{fig:loss-and-network}, it seems that in our setup, smaller networks get bigger gains than bigger ones. Our simple CNN got the highest accuracy around 50\%, followed by DenseNet121 (around 38\%), then Resnet18 (around 31\%) and finally VGG11 (around 20\%). There are no differences on the loss function, which is surprising given that salient loss is dropping all but one of the output dimensions. It suggests that the alignment is a very focused process, that does not require abrupt changes on the latent space.

\subsection{Learning Rate Sensitivity}
\begin{figure}[htbp]
    \centering
    \includegraphics[width=1\linewidth]{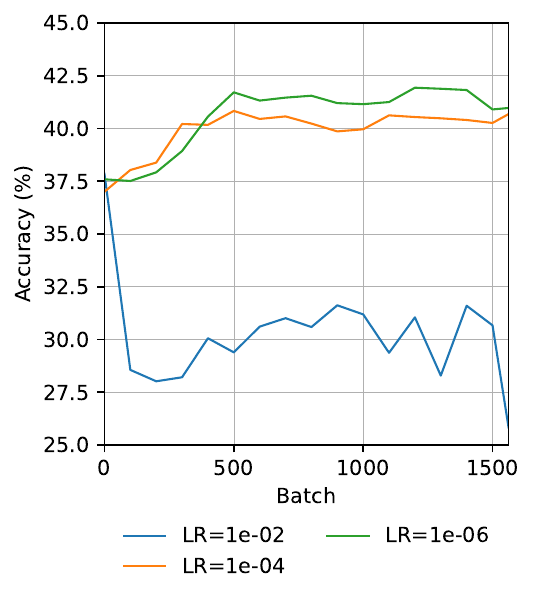}
    \caption{Comparison of KNN (K=5) accuracy varying the learning rate.}
    \label{fig:learning-rate-knn}
\end{figure}

\begin{figure}[htbp]
    \centering
    \includegraphics[width=1\linewidth]{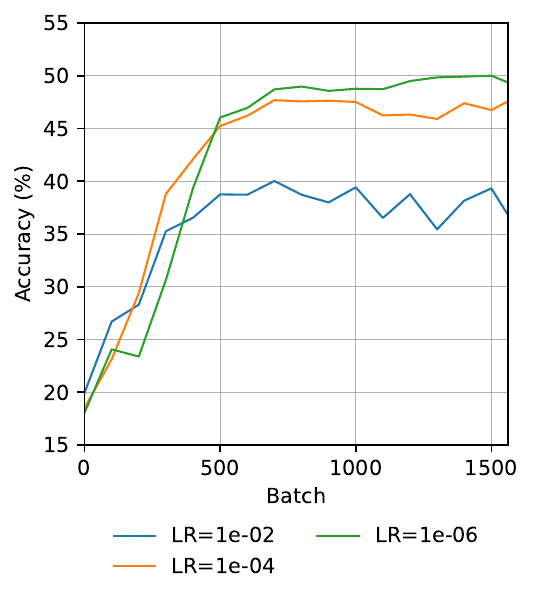}
    \caption{Comparison of linear probe accuracy varying the learning rate.}
    \label{fig:learning-rate}
\end{figure}

As shown in Figure \ref{fig:learning-rate-knn} and Figure \ref{fig:learning-rate}, we found that a lower learning rate produces more useful learned representations, both with KNN or a linear probe. Other interesting aspect to highlight is that the initial representations of the random network are more amenable for KNN classification, meaning that even a randomly initialized network preserves some kind of distances related with actual data features.

\section{Discussion}\label{sec:Discussion}
In our setup, we found that it is possible to get non-trivial accuracy gains from just self-distilling knowledge between peers in a group of randomly initialized networks.

\begin{figure}[htbp]
    \centering
    \includegraphics[width=1\linewidth]{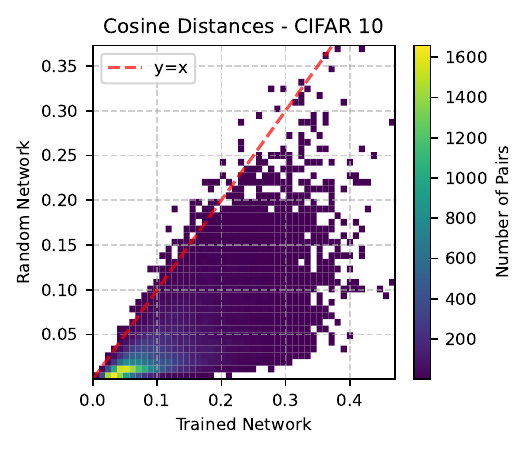}
    \caption{Comparison of the cosine distances between the CIFAR-10 embeddings generated by the untrained network and the ones produced by the trained network.}
    \label{fig:distance-shift}
\end{figure}

To further investigate what is being learned, we compared the distances between samples as measured by the initial untrained neural network and by the trained network. For this, we took CIFAR-10 images and compared its embeddings with a shuffled list of the same embeddings, both for the untrained network and the trained one (keeping the same shuffling). As shown in Figure \ref{fig:distance-shift}, we observe a global shift toward increased separation, particularly among initially close samples. This emergent expansion behavior may partially explain why some non-contrastive self-supervised methods, such as BYOL and DINO, can succeed even in the absence of explicit negative samples. However, a full understanding of this mechanism and its relationship to semantic learning will require further study.

\section{Conclusions and Future Work}\label{sec:Conclusion}
In this work, we studied the effect of self-distillation on learning. Our findings suggest that self-distillation alone can yield improved representations. However, these effects are variable, depending on the network architecture or the learning rate.

Interestingly, we found that the distances in the learned representations are usually bigger than in the initial network. The nature of this expansion needs further analysis, which we leave for future work.

\bibliographystyle{./IEEEtran}
\bibliography{bibliografia}

\end{document}